\documentclass{article}

\usepackage[final,nonatbib]{neurips_2022}

\usepackage[utf8]{inputenc} 
\usepackage[T1]{fontenc}    
\usepackage{hyperref}       
\usepackage{url}            
\usepackage{booktabs}       
\usepackage{amsfonts}       
\usepackage{nicefrac}       
\usepackage{microtype}      
\usepackage[table]{xcolor}         
\usepackage{graphicx}
\usepackage{amsthm}
\usepackage{amsfonts}
\usepackage{amsmath}
\usepackage{wrapfig}
 
\usepackage{pifont}

\usepackage{mathtools}

\usepackage{booktabs}
\usepackage{multirow}
\usepackage{siunitx}
\usepackage{amssymb}
\usepackage{stackengine}

\usepackage{caption}
\usepackage{subcaption}
\usepackage{tikz-cd}
\usepackage{enumitem}

\usepackage[linesnumbered,ruled]{algorithm2e}

\theoremstyle{definition}

\newcommand{\Exp}{\text{Exp}}
\title{Regression-Based Elastic Metric Learning on Shape Spaces of Cell Curves}

\author{
  Adele Myers \\
  University of California, Santa Barbara\\
  Santa Barbara, CA 93106 \\
  \texttt{adele@ucsb.edu} \\
   \And
   Nina Miolane \\
   University of California, Santa Barbara \\
   Santa Barbara, CA 93106 \\
   \texttt{ninamiolane@ucsb.edu} \\
}

\usepackage{biblatex}
\begin{document}
\maketitle
\begin{abstract}
    We propose a metric learning paradigm, Regression-based Elastic Metric Learning (REML), which optimizes the elastic metric for geodesic regression on the manifold of discrete curves. Geodesic regression is most accurate when the chosen metric models the data trajectory close to a geodesic on the discrete curve manifold. When tested on cell shape trajectories, regression with REML's learned metric has better predictive power than with the conventionally used square-root-velocity (SRV) metric. The code is publicly available \href{https://github.com/bioshape-lab/dyn}{here}. 
\end{abstract}

\section{Introduction}

Cell shape is strongly representative of cell function and can help diagnose many conditions including cancer \cite{Prasad2019}. However, most cell analyses do not consider the cell's shape in its entirety and instead only consider coarse global attributes like area, perimeter, or convexity \cite{Gabbert2021-vr}. Analyses that do use the cell's entire shape are often ``static'', as they disregard cell shape as it evolves over time \cite{zernike2016,Phillip_2021}. Because advances in live imaging are poised to provide an increasing amount of cell video data, statistical tools that describe the time evolution of cell shape are timely and necessary to precisely assess cell health and potential pathological conditions.

\begin{figure}[!ht]
    \centering
    \includegraphics[width=\textwidth]{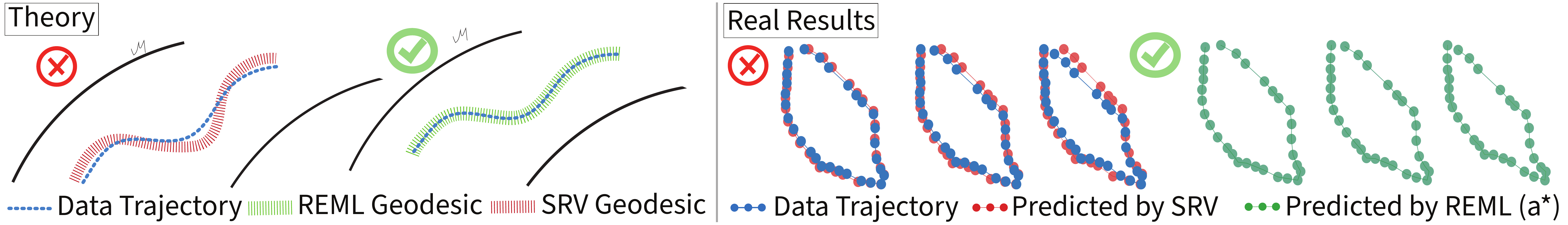}
    \caption{Summary and results of the proposed approach. Left: A trajectory may follow a geodesic as calculated by one metric but not follow a geodesic as calculated by another metric. Our paradigm learns the elastic metric (parameterized by $a$) that best represents the data trajectory as a geodesic on the manifold of discrete curves. Right: true cell trajectory overlaid with 1) cells predicted by a regression which utilizes our paradigm's learned metric parameter ($a*$) 2) cells predicted by a square-root-velocity (SRV) regression. Regression predictions using the SRV metric (red) do not match the data trajectory (blue), but our algorithm's $a*$ predicts the data trajectory perfectly: our prediction (green) perfectly overlays the data trajectory (blue).}
    \label{fig:summary}
\end{figure}

We propose a \textit{metric learning} paradigm, Regression-based Elastic Metric Learning (REML). REML learns the optimal elastic metric for geodesic regression on the manifold of discrete curves: here, the manifold of cell shapes. Our work expands on the framework \cite{Bharath2020-zn,Miolane2022ARiemannian} which parameterizes cell shapes from microscopy images with a 2D array of discrete points that trace a cell's outline. Then, we use the elastic metric implemented in Geomstats \cite{miolane2020geomstats} to analyze these shapes on the manifold of discrete curves. The outline of each cell is itself a point on the manifold of discrete curves $\mathcal{M}$, and changes in cell shape over time form a trajectory on $\mathcal{M}$. Our method analytically solves the geodesic regression problem using different metrics and selects the metric that maximizes the coefficient of determination $R^2$ on the validation set Fig.~\ref{fig:summary}. The elastic metric is particularly meaningful when analyzing cell shape because it quantifies ``stretching'' and ``bending'' between curves, which provides a biological measure of stretching and bending properties of the cells in a cell shape trajectory. We validate our approach on synthetic trajectories between real osteosarcoma cells. The experimental results confirm that our paradigm (i) learns biological parameters of cell shape evolution, and (ii) provides a mathematically-grounded approach to enhanced characterization of cell shape evolution.

\section{Background}

This section reviews the tools of Riemannian geometry that will support our metric learning method. Additional background can be found in \cite{needham2020simplifying,bauer2022elastic}. 
We model cell outlines as elements of the space of regular planar curves $\mathcal{C}=\mathcal{C}^1([0, 1], \mathbb{R}^2)$. Our metric learning approach will require us to calculate distances between cell shapes in $\mathcal{C}$. To do so, we introduce the concept of elastic metrics. 

\paragraph{Elastic metrics} A \textit{Riemannian metric} $g$ on a manifold $\mathcal{C}$ is a set of smoothly varying inner products defined on tangent spaces of $\mathcal{C}$. These define geometric measurements on $\mathcal{C}$, including distances, geodesics and exponential maps. Geodesics on manifolds generalize straight lines on vector spaces: they are curves that locally minimize the distance between points. Exponential maps generalize addition on vector spaces.



We equip $\mathcal{C}$ with a family of \textit{elastic metrics $ g^{a, b}$} parameterized by $a, b > 0$ \cite{mio2007shape}:
\begin{equation}\label{eq:elmetric}
 g^{a, b}_c(h, k) = a^2\int_{0}^1\langle D_sh, N\rangle\langle D_sk, N\rangle ds + b^2 \int_{0}^1\langle D_sh, T\rangle\langle D_sk, T\rangle ds, \enspace \forall c \in \mathcal{C}, h, k \in T_c\mathcal{C},
\end{equation}
where $c$ is a cell outline (a point in $\mathcal{C}$), $h, k$ are infinitesimal deformations of $c$ (tangent vectors in $T_c\mathcal{C}$), $D_s$ is the derivative with respect to arc-length $s$ along the curve $c$, $(T,N)$ denote the unit tangent and the unit normal to $c$, respectively, and $<,>$ is the canonical Euclidean inner-product of $\mathbb{R}^2$. An elastic metric $ g^{a, b}$ depends on two parameters: a ``bending'' parameter $b$ and a ``stretching'' parameter $a$, which respectively evaluate whether two curves are ``bent'' or ``stretched'' compared to one another and then define how far apart these curves should be on $\mathcal{C}$. For example, when $a$ is large, two curves that differ by a stretching operation will be far apart. When $b$ is large, two curves that differ by a bending operation will be far apart. Elastic metrics are invariant under shape-preserving transformations, i.e., elements in the group $\mathcal{G}$ of 2D translations, 2D rotations, and re-scaling of cell outlines \cite{needham2020simplifying}. As such, elastic metrics are well-defined on the \textit{space of curve shapes}, which is formally given by the quotient $\mathcal{M} = \mathcal{C}^1([0, 1], \mathbb{R}^2) / \mathcal{G}$. The elastic metric with $a=1$ and $b=0.5$ is called the square-root-velocity (SRV) metric, and is often used as the “default metric” on the manifold of discrete curves. For this reason, we will consider the statistical analysis with the SRV metric as a baseline in our experiments.

\paragraph{Geodesic regression} Because $\mathcal{M}$ is a manifold, we consider geodesic regression, which is the linear regression equivalent for manifolds. \textit{Geodesic regression} on $(\mathcal{M}, g^{a, b})$ solves a least-square fitting problem \cite{Fletcher2011}:
\begin{equation}\label{eq:geo-regression}
    \min_{(p, v) \in T\mathcal{M}} \sum_{i=1}^{T} d^2 \left(\Exp_p(t_i v), c_i \right),
\end{equation}
where $d$ and $\Exp$ are the Riemannian distance and exponential maps associated with the metric $g^{a, b}$. When the metric is Euclidean, this expression simplifies to the usual linear regression with intercept $p$ and slope $v$.
In geodesic regression, choice of metric affects the goodness of fit. Because metrics define distances between points in shape space, a data trajectory may follow the geodesic calculated by one metric but not another. We design an algorithm that finds the "optimal" metric for regression of a particular trajectory \textemdash the metric where the trajectory is closest to a geodesic, as judged by the value of the coefficient of determination $R^2$. This optimal metric gives the regression fit more predictive power, which in the case of microscopy image analysis in cell biology can thus provide a more accurate prediction of future cell shapes. We will perform metric learning across all possible elastic metrics and compare our results to the commonly used SRV metric.

\paragraph{Related Works} (i) \textit{Shape Spaces:} Cell biology has only recently started to use tools of statistical shape analysis. Like us, Philip \textit{et al.} study cell shapes, but unlike us, they use the Kendall metric, which is a metric designed to compare shapes defined as sets of landmarks, while the elastic metric is meant to study shapes of curves \cite{Bharath2020-zn}. Cho \textit{et al.} use the SRV metric to study biological shapes, but they do not generalize their study to all elastic metrics, and they do not study cell shape as we do \cite{Cho2019-vi}. (ii) \textit{Metric Learning:} The problem of metric learning has received considerable attention in machine learning literature for many years (see \cite{kulis2013metric} for a survey). However, no existing paradigm optimizes the elastic metric for geodesic regression on cell shape trajectories. Riemannian metric learning has been studied by \cite{lebanon2012learning,sauty2022riemannian}, but these works do not learn elastic metrics. In research concurrent to ours, Bauer \textit{et al.} \cite{bauer2022elastic} proposed an \textit{elastic metric learning} scheme. However, they optimize the elastic metric in the context of shape classification, while we optimize it here for geodesic regression. Additionally, they use a random search on the elastic metrics' parameter, whereas we propose a gradient-ascent scheme instead.

\section{Methods}

In what follows, we consider synthetic trajectories of osteosarcoma cell shapes changing through time. The synthetic trajectory draws a geodesic between two real osteosarcoma cells, and thus, the synthetic shapes in the trajectory mimic realistic cell outlines. We give each trajectory a predetermined (i) number of time points, (ii) number of sampling points on each cell outline, and (iii) amount of measurement noise on each sampling point. These discrete cell outlines lie on the manifold of discrete curves, which we equip with an elastic metric. We divide the time trajectory of cell shapes $c_1, ..., c_T$ into three sequential datasets: a train set (60\%), a validation set (30\%) and a test set (10\%), the size of which depends on the size of the synthetic trajectory. Our paradigm uses the train set to learn regression fit parameters for any metric $g_{ab}$ being tested by the validation set. The validation set allows us to learn the ''optimal'' metric, chosen as the metric that maximizes the value of the coefficient of determination $R^2$. We denote $a*$ the parameter of this optimal elastic metric. We use the test data set to evaluate the regression performance of the learned $a*$ to that of the SRV metric.

\paragraph{1. Training: Geodesic regression with $F_{a,b}$-transform} Our training procedure simplifies regression fit calculations by harnessing a key property of elastic metric $g^{a, b}$. Riemannian operations on cell shapes $c$ on the manifold $(\mathcal{M}, g^{a, b})$ are equivalent to Euclidean operations in a linear ``$F_{ab}$ space'', which is accessible through the $F_{ab}$ transform defined as \cite{needham2020simplifying}: 
\begin{equation}
    F_{ab}(c) = 2b\|c'\|^{1/2} \left(\frac{c'}{\|c'\|}\right)^{\frac{a}{2b}},
\end{equation}
where $c'$ is the derivative of the curve w.r.t. its parameter $s\in [0, 1]$. The $F_{ab}$-transform maps a curve $c$, which lies on the Riemannian manifold $\mathcal{M}$, to a point in the Euclidean $F_{ab}$-space. The $F_{ab}$-transform also has an inverse, which maps elements of the $F_{ab}$-space back to curve shapes on $\mathcal{M}$. This allows us to perform Euclidean operations in $F_{ab}$ space and then transform the results back into shape space. Using the $F_{ab}$-transform, the geodesic regression of Eq.~\ref{eq:geo-regression} becomes a linear regression on the train set $F_{ab}(c_1), ..., F_{ab}(c_{T_{\text{train}}})$. 


\paragraph{2. Validation: Optimization of the elastic parameter $a$} We use the validation set to perform elastic metric learning. The definition of the elastic metric in Eq.~\ref{eq:elmetric} shows that the parameter $b$ is only a rescaling factor, as only the ratio $\frac{a}{b}$ substantially affects the metric. Different choices of $b$ only change units and do not affect whether a trajectory is a geodesic or not. 
Thus, we fix the "units of the calculation" by imposing $b=0.5$ and perform elastic metric learning by optimizing only $a$. We evaluate how close a trajectory is to a geodesic by calculating the $R^2$ value of the fitted geodesic regression model:
\begin{equation}
 R^2 = 1- \frac{\sum_{i=i_{train}}^{i_{val}}\|y_i -\hat{y_i} \|}{\sum_{i_{train}}^{i_{val}}\|y_i -\bar{y} \|},
\end{equation}
where $y_i = F_{a}(c_i)$ and $\bar{y} = \frac{1}{n_{\text{val}}} \sum_{i=i_{train}}^{i_{val}} y_i$.  $\hat y_i$ is the $i^{th}$ cell shape predicted by the regression model parameters derived from the "training" data set. Note that $R^2$ depends explicitly on the elastic parameter $a$, which allows us to compute its gradient with respect to $a$ (see analytical expression in the appendices). 

We perform gradient ascent on the coefficient of determination $R^2$ of the validation set to find the optimal $a$, the $a*$ that brings $R^2$ closest to 1. Importantly, we derive the gradient of $R^2$ with respect to $a$ (see appendices for derivation).

\paragraph{3. Evaluation and Test} We evaluate our metric learning method based on 1) whether the $a*$ calculated by the validation dataset is close to $a_{true}$ (which is the metric used to create the synthetic data trajectory), 2) whether regression on the test dataset using $a*$ provides a good fit and $R^2$ value, and 3) whether the regression fit on the test dataset using $a*$ is better or worse than the regression fit using the SRV metric.

\section{Experiments}

\paragraph{Datasets} We simulate the evolution of a cancer cell through time by generating an extensive set of synthetic geodesics between two cancer cells extracted from real microscopy images \cite{Miolane2022ARiemannian}. We generate geodesics with all combinations of the following parameters: $a_{true} \in \{0.2, 0.5, 0.7, 0.9, 1.0\}$ ($a$ used to generate synthetic geodesic trajectory i.e., the metric space where the trajectory follows a geodesic), number of time points $T$ (number of cells on the geodesic): $T \in \{20, 50, 100, 200\}$, number of sampling points $n_s$ along each cell outline $n_s \in \{30, 50, 100\}$, and standard deviation of a centered Gaussian noise $\sigma \in \{0.0, 0.001, 0.01\}$ added on the 2D coordinates of the sampling points. Our synthetic experiments are directly relevant to biological researchers as they help inform the characteristics of their experimental cell shape dataset. The number of time points $T$ corresponds to a choice of length of biological experiment video recording; the number of sampling points $n_s$ is a biologist's choice upon extracting a discrete cell outline from a microscopy image, and the noise level $\sigma$ represents the uncertainty during the cell outline extraction process from images.

\begin{figure}[h!]
    \centering
    \includegraphics[width=\textwidth]{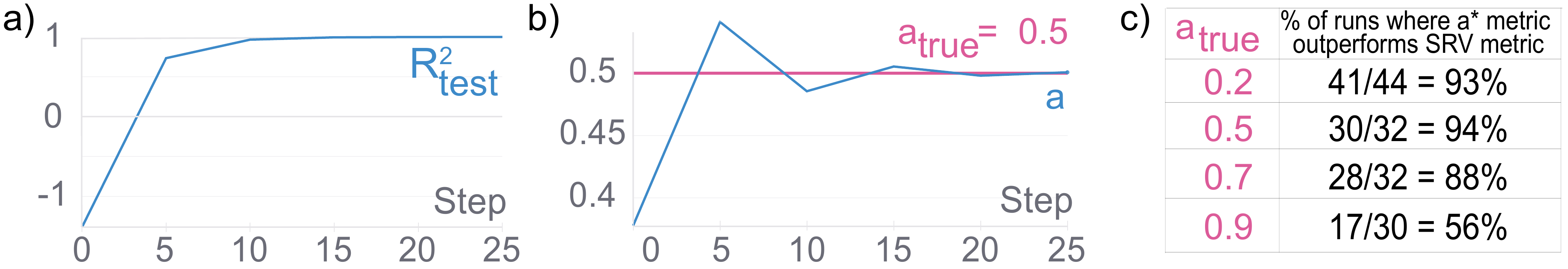}
    \caption{Experimental results on synthetic trajectories modeling the evolution of real osteosarcoma cells. (a-b) shows our method simultaneously converging upon $R^2 = 1$ and $a_{true} = 0.5$. c) When the number of cell sampling points is between $30-50$, the number of cells in the trajectory is $100-200$, and the noise level is $<0.005$, the $a*$ metric predicted by our algorithm outperforms the SRV metric with a better $R^2$ at every $a_{true}$ (excluding $a_{true}=1$, which is the SRV metric itself).}
    \label{fig:results}
\end{figure}

\paragraph{Results} Our experimental results demonstrate that (i) our algorithm successfully converges $a*$ to $a_{true}$ and $R^2_{test}$ to $1$ and (ii) our method outperforms the SRV metric at every $a_{true}$ excluding $a_{true}=SRV$ when the noise level is $<0.005$, the cells are sampled between $30-50$ times, and there are $100-200$ cells in the trajectory. Our results show that our method (i) correctly learns the metric $a_{true}$ that was used to generate the synthetic cell trajectory and (ii) provides better predictions of future cell shapes. Learning $a$ is biologically relevant because $a$ quantifies how much a cell bends and stretches over a given amount of time.

\section{Conclusion and Future Work}
\label{sec:conclusion}
Our method, Regression-based Elastic Metric learning, describes an automated way for finding the elastic metric that brings a trajectory closest to a geodesic on the manifold of discrete curves. Our method increases the predictive power of geodesic regression, which allows for more accurate prediction of future cell shapes. In the long term, such methodology could be used for early detection of diseases or for early diagnosis of chemo-resistant  cells based on their early evolution. Currently, most analyses on the manifold of discrete curves use the SRV metric by default. However, our results show that the SRV metric does not produce the best regression fit for most trajectories. Our method recognizes that the "optimal" metric is trajectory-dependent and thus creates an opportunity for improved analysis on all cell data trajectories.


\newpage
\printbibliography

\newpage
\section{Appendix}
\subsection{Derivation of $R^2$ in terms of $a$} This derivation finds $R^2$ in terms of $a$. We are considering regression on curve trajectories, so $R^2$ is given by 

\begin{equation}
 R^2 = 1- \frac{\sum_{n_{train}}^{n_{val}}d(c_i,\hat{c_i} )^2}{\sum_{n_{train}}^{n_{val}}d(c_i ,\bar{c} )^2} 
\end{equation}.

where $c_i$ is an arbitrary curve in the trajectory, $\hat{c_i}$ is a curve predicted by the regression model, $\bar{c}$ is the mean curve in the data set curve trajectory, $n_{train}$ is the number of curves in the data set trajectory that are used to train the regression model, and $n_{val}-n_{train}$ is the number of curves in the validation data set trajectory that we use to calculate $R^2$ (and thus, the number of curves that we use to optimize $R^2$ to find $a*$). The numerator of the second term is the (un-normalized) mean squared error, which shows how much variation exists between the data set curve trajectory and the regression-predicted trajectory. The denominator of the second term is the (un-normalized) variance, which shows how much variation exists between the curves in the data set trajectory. 

\paragraph{We will begin by finding the mean squared error (MSE) in terms of $a$.}

\begin{align*}
    MSE 
    &= \sum_{i=n_{train}}^{n_{val}} \| \Vec{r_i} \|^2 \\
    &= \sum_{i=n_{train}}^{n_{val}} d( c_i, \hat c_i)^2.
\end{align*}

We choose to do this analysis in $F_a-space$, and use $q$'s to denote curves which have been transformed to $F_a-space$. Thus, the derivation continues as

\begin{align*}
    &= \sum_{i=n_{train}}^{n_{val}} \| q_i - \hat q_i \|^2 \\
    &= \sum_{i=n_{train}}^{n_{val}} \| q_i - \hat \beta_0 - \hat \beta_1 t_i \|^2 \\
\end{align*}

\paragraph{To continue our derivation of the MSE in terms of $a$, we must derive expressions for the $\beta$ parameters in terms of $a$.}

Consider the linear regression in F-space:
\begin{equation}
    q_i = \beta_0 + \beta_1 t_i.
\end{equation}
We will learn/estimate the coefficients $\beta_0, \beta_1$ using the training set of cells. We will learn the optimal $a$ factor using the validation set. The size of the training set is denoted $n_{train}$, the size of the validation set is denoted $n_{val}-n_{train}$.

The estimated coefficients of the regressions are given by the normal equations, i.e.
\begin{equation}
    \hat \beta = (X^TX)^{-1}X^T y
\end{equation}
where $\hat \beta = \begin{bmatrix} \hat \beta_0 \\ \hat \beta_1 \end{bmatrix}$ are the estimated coefficients, $X = \begin{bmatrix} 1 & t_1 \\ \vdots & \vdots \\ 1 & t_n  \end{bmatrix}$ is the design matrix, $y = \begin{bmatrix} y_1 & \dots & y_n \end{bmatrix}$ is the response.
We have
\begin{equation}
 X^TX = \begin{bmatrix} n_{train} & \sum_{i=1}^{n_{train}} t_i \\ \sum_{i=1}^{n_{train}} t_i & \sum_{i=1}^{n_{train}} t_i^2\end{bmatrix} = {n_{train}} \begin{bmatrix} 1 & \bar t \\ \bar t & \bar t^2\end{bmatrix},   
\end{equation}
where we have introduced the notations: $\bar t = \sum_{i=1}^{n_{train}} t_i$ and $\bar t^2 = \sum_{i=1}^{n_{train}} t_i^2$.

The inverse gives:
\begin{equation}
 (X^TX)^{-1} = \frac{1}{{n_{train}} (\bar t^2 - (\bar t)^2)} \begin{bmatrix} t^2 & -\bar t \\ -\bar t & \bar 1\end{bmatrix}.  
\end{equation}

Multiplying by $X^T$:
\begin{equation}
 (X^TX)^{-1}X^T = \frac{1}{{n_{train}} (\bar t^2 - (\bar t)^2)} \begin{bmatrix} 
 \bar t^2 - \bar t . t_1 & \hdots & \bar t^2 - \bar t . t_{n_{train}}\\ 
 -\bar t + t_1 & \hdots &  -\bar t + t_{n_{train}}
 \end{bmatrix}.  
\end{equation}

Multiplying by $y$:
\begin{equation}
 \hat \beta 
 = (X^TX)^{-1}X^T y 
 = \frac{1}{{n_{train}} (\bar t^2 - (\bar t)^2)} 
 \begin{bmatrix} 
 \sum_{i=1}^{n_{train}} (\bar t^2 - \bar t . t_i)q_i\\ 
 \sum_{i=1}^{n_{train}} ( -\bar t + t_i)q_i
 \end{bmatrix}.  
\end{equation}

We introduce the notations:
\begin{equation}
    \tau_{i0} = \frac{(\bar t^2 - \bar t . t_i)}{{n_{train}} (\bar t^2 - (\bar t)^2)}; \qquad \tau_{i1} = \frac{( -\bar t + t_i)}{{n_{train}} (\bar t^2 - (\bar t)^2)}
\end{equation}
so that:
\begin{equation}
    \hat \beta 
    =  \begin{bmatrix} 
 \sum_{i=1}^{n_{train}} \tau_{i0} q_i\\ 
 \sum_{i=1}^{n_{train}} \tau_{i1} q_i
 \end{bmatrix}
    = \begin{bmatrix} 
 \sum_{i=1}^{n_{train}} \tau_{i0} F_{ab}(c_i)\\ 
 \sum_{i=1}^{n_{train}} \tau_{i1} F_{ab}(c_i)
 \end{bmatrix},
\end{equation}
where the last expression highlights the dependency of this quantity in a, b.

\paragraph{Thus, we can continue our MSE derivation as:}

\begin{align*}
    &= \sum_{i=n_{train}}^{n_{val}} 
        \| q_i 
        - \sum_{j=1}^n \tau_{j0} q_j
        - \sum_{j=1}^n \tau_{j1} q_j t_i \|^2 \\
    &= \sum_{i=n_{train}}^{n_{val}} 
        \| q_i 
        - \sum_{j=1}^n (\tau_{j0} + \tau_{j1} t_i) q_j\|^2  \\
    &= \sum_{i=n_{train}}^{n_{val}} 
        \| F_{ab}(c_i) 
        - \sum_{j=1}^n (\tau_{j0} + \tau_{j1} t_i) F_{ab}(c_j)\|^2  \\  
    &= \sum_{i=n_{train}}^{n_{val}}
        \| F_{ab}(c_i) 
        - \sum_{j=1}^n \tau_{ij} F_{ab}(c_j)\|^2,  \\ 
\end{align*}

where we denote $n = n_{train}$ for convenience of notations.

\paragraph{Now, let's begin our derivation of the (un-normalized) variance (VAR) of the data set curve trajectory.}

\begin{align*}
    VAR = 
    &= \sum_{i=n_{train}}^{n_{val}} d( c_i , \bar{c} )^2 \\
     &= \sum_{i=n_{train}}^{n_{val}}
        \| F_{ab}(c_i) 
        - \frac{1}{n_{val}}\sum_{j=1}^{n_{train}} F_{ab}(c_j)\|^2,  \\ 
\end{align*}

where the "f-transform" explicitly depends on $a$.

\paragraph{Thus, in its entirety, $R^2$ w.r.t. $a$ is given by:}

\begin{equation}
    R^2 = 1- \frac{\sum_{i=n_{train}}^{n_{val}}
        \| F_{ab}(c_i) 
        - \sum_{j=1}^n \tau_{ij} F_{ab}(c_j)\|^2}{\sum_{i=n_{train}}^{n_{val}}
        \| F_{ab}(c_i) 
        - \frac{1}{n_{val}}\sum_{j=1}^{n_{train}} F_{ab}(c_j)\|^2}
\end{equation}

The next section expands the norm in this result.

\subsection{Expansion of the norm $\| \Vec{r_i} \|^2$.} The previous section derives $R^2$ in terms of $a$, and comes to the result:

\begin{equation}
    R^2 = 1- \frac{\sum_{i=n_{train}}^{n_{val}}
        \| F_{ab}(c_i) 
        - \sum_{j=1}^n \tau_{ij} F_{ab}(c_j)\|^2}{\sum_{i=n_{train}}^{n_{val}}
        \| F_{ab}(c_i) 
        - \frac{1}{n_{val}}\sum_{j=1}^{n_{train}} F_{ab}(c_j)\|^2}
\end{equation}

This section will show how we expand the norm in the equation above. We will use the MSE norm as an example. $\Vec{r_i}$ is a vector which indicates the difference between two curves: $c_i$, which is the $i^{th}$ curve in the data set trajectory, and $\hat{c_i}$, which is the corresponding $i^{th}$ curve in the regression-predicted trajectory. Because $\Vec{r_i}$ indicates a difference between two curves, it must be a function of the parameter $s$, which parameterized all curves. Thus, its long-form notation is $\Vec{r_i(s)}$.  Conceptually, if all curves are deformed circles, then the parameter $s$ tells you where a point on a curve maps to on an equivalent circle. $s$ ranges from $0$ to $1$ by construction, where $s=0$ is the start point of the curve and $s=1$ is the end point of the curve. 
The squared vector norm of the function $\Vec{r_i(s)}$ is given by

\begin{align*}
     \sum_{i=n_{train}}^{n_{val}} \| \Vec{r_i} \|^2 = \sum_{i=n_{train}}^{n_{val}} \int_{0}^{1} \| \Vec{r_i(s)} \|^2 \,ds 
     &= \sum_{i=n_{train}}^{n_{val}} \int_{0}^{1} \Vec{r_i(s)} \cdot \Vec{r_i(s)} \,ds
\end{align*}

Now, we will describe how we compute this integral. In our code, we take discrete samples of the points in each curve. Therefore, in our code, a curve is represented by a 2D array containing the $(x,y)$ coordinates of these samples. Because $c_i$'s in curve space are 2D discrete curves, $q_i$'s are also discrete curves (in the linear $F_a-space$), and $\Vec{r_i}$'s are discrete vectors comprised of the coordinate differences between two curves. Because $\Vec{r_i}$'s are discrete, we will compute the integral above using a Riemann sum.

\begin{align*}
     \sum_{i=n_{train}}^{n_{val}} \| \Vec{r_i} \|^2 
     &= \sum_{i=n_{train}}^{n_{val}} \sum_{s=0}^{1} \Vec{r_i(s)} \cdot \Vec{r_i(s)} \,\Delta s
\end{align*}

Here, $\Delta s = \frac{\text{1ength of curve}}{\text{number of segments}}=\frac{1}{\text{number of sampling points} -1}$ since in $F_a-space$, the transformed curve $q_i$ has one fewer sampling point than $c_i$.

\subsection{Derivation of $R^2$'s derivative with respect to a} 
In the previous appendix sections, we derived $R^2$ w.r.t. $a$. Now, we will compute the gradient of $R^2$ w.r.t. $a$.

\begin{align*}
    R^2 = 1- \frac{\sum_{i=n_{train}}^{n_{val}}
        \| F_{ab}(c_i) 
        - \sum_{j=1}^n \tau_{ij} F_{ab}(c_j)\|^2}{\sum_{i=n_{train}}^{n_{val}}
        \| F_{ab}(c_i) 
        - \frac{1}{n_{val}}\sum_{j=1}^{n_{train}} F_{ab}(c_j)\|^2}
\end{align*}
    
\begin{align*}
\frac{d}{da}R^2 
    &= -\frac{d}{da} \frac{\sum_{i=n_{train}}^{n_{val}}
    \| F_{ab}(c_i) 
    - \sum_{j=1}^n \tau_{ij} F_{ab}(c_j)\|^2}{\sum_{i=n_{train}}^{n_{val}}
    \| F_{ab}(c_i) 
    - \frac{1}{n_{val}}\sum_{j=1}^{n_{train}} F_{ab}(c_j)\|^2} \\
    &= -\frac{d}{da}\frac{MSE}{VAR}\\
    &= -\frac{\frac{d}{da}MSE * VAR - MSE* \frac{d}{da}VAR}{VAR^2}
\end{align*}

We now calculate $\frac{d}{da}MSE$ and $\frac{d}{da}VAR$.

\paragraph{Let's start with $\frac{d}{da}MSE$.} Using information from the previous section, we see

\begin{align*}
    \frac{d}{da}MSE 
    &= \frac{d}{da}\sum_{i=n_{train}}^{n_{val}} \sum_{s=0}^{1} \Vec{r_i(s)} \cdot \Vec{r_i(s)} \,\Delta s \\
     &= 2 \sum_{i=n_{train}}^{n_{val}} \sum_{s=0}^{1} \left[\frac{d}{da}\Vec{r_i(s)}\right] \cdot \Vec{r_i(s)} \,\Delta s \\
\end{align*}

where for the MSE, $r_i(s) = F_{ab}(c_i(s)) - \sum_{j=1}^n \tau_{ij} F_{ab}(c_j(s))$. Calculating $\frac{d}{da} \Vec{r_i(s)}$ is the trickiest part. Halfway through the calculation, we switch into polar coordinates because polar coordinates allow us to easily take derivatives of vectors raised to the power of the differentiated variable (in this case, $a$).

\begin{align*}
    \frac{d}{da} \Vec{r_i(s)}
    &= \frac{d}{da}\left[|c_i'|^{1/2}\frac{c_i'}{|c_i'|}^a- \sum_{j=1}^n \tau_{ij}|c_j'|^{1/2}\frac{c_j'}{|c_j'|}^a\right]\\
    &= |c_i'|^{1/2}\frac{d}{da}\frac{c_i'}{|c_i'|}^a- \sum_{j=1}^n \tau_{ij}|c_j'|^{1/2}\frac{d}{da}\frac{c_j'}{|c_j'|}^a\\
    &= |c_i'|^{1/2}\frac{d}{da}e^{i arg(c_i')a}- \sum_{j=1}^n \tau_{ij}|c_j'|^{1/2}\frac{d}{da}e^{i arg(c_j')a}\\
    &= |c_i'|^{1/2}(i arg(c_i'))e^{i arg(c_i')a}- \sum_{j=1}^n \tau_{ij}|c_j'|^{1/2}(i arg(c_j'))e^{i arg(c_j')a}\\
    &= |c_i'|^{1/2}(e^{i \pi/2} arg(c_i'))e^{i arg(c_i')a}- \sum_{j=1}^n \tau_{ij}|c_j'|^{1/2}(e^{i \pi/2} arg(c_j'))e^{i arg(c_j')a}\\
    &= |c_i'|^{1/2}(arg(c_i'))e^{i( \pi/2+ arg(c_i')a)}- \sum_{j=1}^n \tau_{ij}|c_j'|^{1/2}(arg(c_j'))e^{i( \pi/2+ arg(c_j')a)}\\
\end{align*}

where we have used the facts that $e^{i arg(c_i')}= \frac{c_i'}{|c_i|}$ and $i = e^{i \pi/2}$. This result can now be plugged into the equation for $ \frac{d}{da}MSE$.

\paragraph{The calculation for $\frac{d}{da}VAR$} is similar, except for the VAR, $r_i(s) = F_{ab}(c_i) 
    - \frac{1}{n_{val}}\sum_{j=1}^{n_{train}} F_{ab}(c_j)$.

\newpage

\section*{Checklist}


\begin{enumerate}

\item For all authors...
\begin{enumerate}
  \item Do the main claims made in the abstract and introduction accurately reflect the paper's contributions and scope?
    \answerYes{}
  \item Did you describe the limitations of your work?
    \answerYes{}
  \item Did you discuss any potential negative societal impacts of your work?
    \answerNA{This work instead encourages ethically responsible computing as it focuses on differential privacy. We thereby do not see any potential for negative social impact.}
  \item Have you read the ethics review guidelines and ensured that your paper conforms to them?
    \answerYes{}
\end{enumerate}

\item If you are including theoretical results...
\begin{enumerate}
  \item Did you state the full set of assumptions of all theoretical results?
    \answerYes{}
        \item Did you include complete proofs of all theoretical results?
    \answerYes{}
\end{enumerate}

\item If you ran experiments...
\begin{enumerate}
  \item Did you include the code, data, and instructions needed to reproduce the main experimental results (either in the supplemental material or as a URL)?
    \answerYes{Will be part of supplementary material}
  \item Did you specify all the training details (e.g., data splits, hyperparameters, how they were chosen)?
    \answerYes{}
        \item Did you report error bars (e.g., with respect to the random seed after running experiments multiple times)?
    \answerNA{}
        \item Did you include the total amount of compute and the type of resources used (e.g., type of GPUs, internal cluster, or cloud provider)?
    \answerNA{computations done locally on a Macbook Pro}
\end{enumerate}

\item If you are using existing assets (e.g., code, data, models) or curating/releasing new assets...
\begin{enumerate}
  \item If your work uses existing assets, did you cite the creators?
    \answerYes{}
  \item Did you mention the license of the assets?
    \answerYes{}
  \item Did you include any new assets either in the supplemental material or as a URL?
    \answerYes{}
  \item Did you discuss whether and how consent was obtained from people whose data you're using/curating?
    \answerNA{No data was curated. All the datasets were publicly available}
  \item Did you discuss whether the data you are using/curating contains personally identifiable information or offensive content?
    \answerNA{}
\end{enumerate}

\item If you used crowdsourcing or conducted research with human subjects...
\begin{enumerate}
  \item Did you include the full text of instructions given to participants and screenshots, if applicable?
    \answerNA{}
  \item Did you describe any potential participant risks, with links to Institutional Review Board (IRB) approvals, if applicable?
    \answerNA{}
  \item Did you include the estimated hourly wage paid to participants and the total amount spent on participant compensation?
    \answerNA{}
\end{enumerate}

\end{enumerate}

\end{document}